\documentclass[11pt]{article}

\usepackage{booktabs}
\usepackage{makecell}
\usepackage[table]{xcolor}
\usepackage{pifont}
\usepackage{soul}
\usepackage{graphicx}
\usepackage{tabularx}
\usepackage{booktabs}

\newcommand{\cmark}{\ding{51}}
\newcommand{\xmark}{\ding{55}}

\usepackage{multirow}

\usepackage[preprint]{acl}

\usepackage{times}
\usepackage{latexsym}

\usepackage[T1]{fontenc}
\usepackage[utf8]{inputenc}

\usepackage{microtype}

\usepackage{inconsolata}

\usepackage{graphicx}

\title{MMRad-22K: A Structured Multimodal Evidence Dataset for Chest X-ray Report Generation}

\author{
  \textbf{Yichen Zhao\textsuperscript{1}$^\dag$ },
  \textbf{Zelin Peng\textsuperscript{1}$^\dag$ },
  \textbf{Fenghe Tang\textsuperscript{2,3}},
  \textbf{Piao Yang\textsuperscript{4}},
  \textbf{Yu Huang\textsuperscript{1}},
  \textbf{Wei Shen\textsuperscript{1}$^*$}
\\
  \textsuperscript{1} MoE Key Lab of Artificial Intelligence, AI Institute,\\  School of Computer Science, Shanghai Jiao Tong University, Shanghai, China\\
  \textsuperscript{2} School of Biomedical Engineering, Division of Life Sciences and Medicine,\\
University of Science and Technology of China (USTC), Hefei, Anhui 230026, China\\
  \textsuperscript{3} Center for Medical Imaging, Robotics, Analytic Computing \& Learning (MIRACLE),\\
Suzhou Institute for Advanced Research, USTC, Suzhou, Jiangsu 215123, China\\
  \textsuperscript{4} Department of Radiology, The First Affiliated Hospital, \\ Zhejiang University School of Medicine, Hangzhou, Zhejiang, China\\
}

\begin{document}
\maketitle
\begin{abstract}
Chest X-ray (CXR) reporting follows a region-based clinical workflow in which radiologists inspect anatomical regions and integrate localized findings into a final report. However, existing resources for CXR report generation provide these supervision signals in fragmented forms. We introduce MMRad-22K, a dataset that organizes regional textual observations, anatomical grounding coordinates, localized image evidence, and report targets into structured multimodal evidence units for CXR report generation. To motivate this formulation, we first compare different evidence formats for report generation and find that structured multimodal evidence is generally more useful than text-only or bounding box-based evidence. We then adapt a unified LVLM backbone using MMRad-22K and show that adaptation with multimodal evidence outperforms both textual-evidence adaptation and end-to-end adaptation on language and clinically oriented metrics. Under the same evaluation protocol, the adapted model also reaches a performance level comparable to several open-source LVLM references. Together, these results support MMRad-22K as a practical structured multimodal resource for training and evaluating CXR report generation aligned with clinical reading workflows. The code is available at \url{https://github.com/qiuzyc/MMRad/}
\end{abstract}

\section{Introduction}

\begin{figure}[t]
    \centering
    \includegraphics[width=\columnwidth]{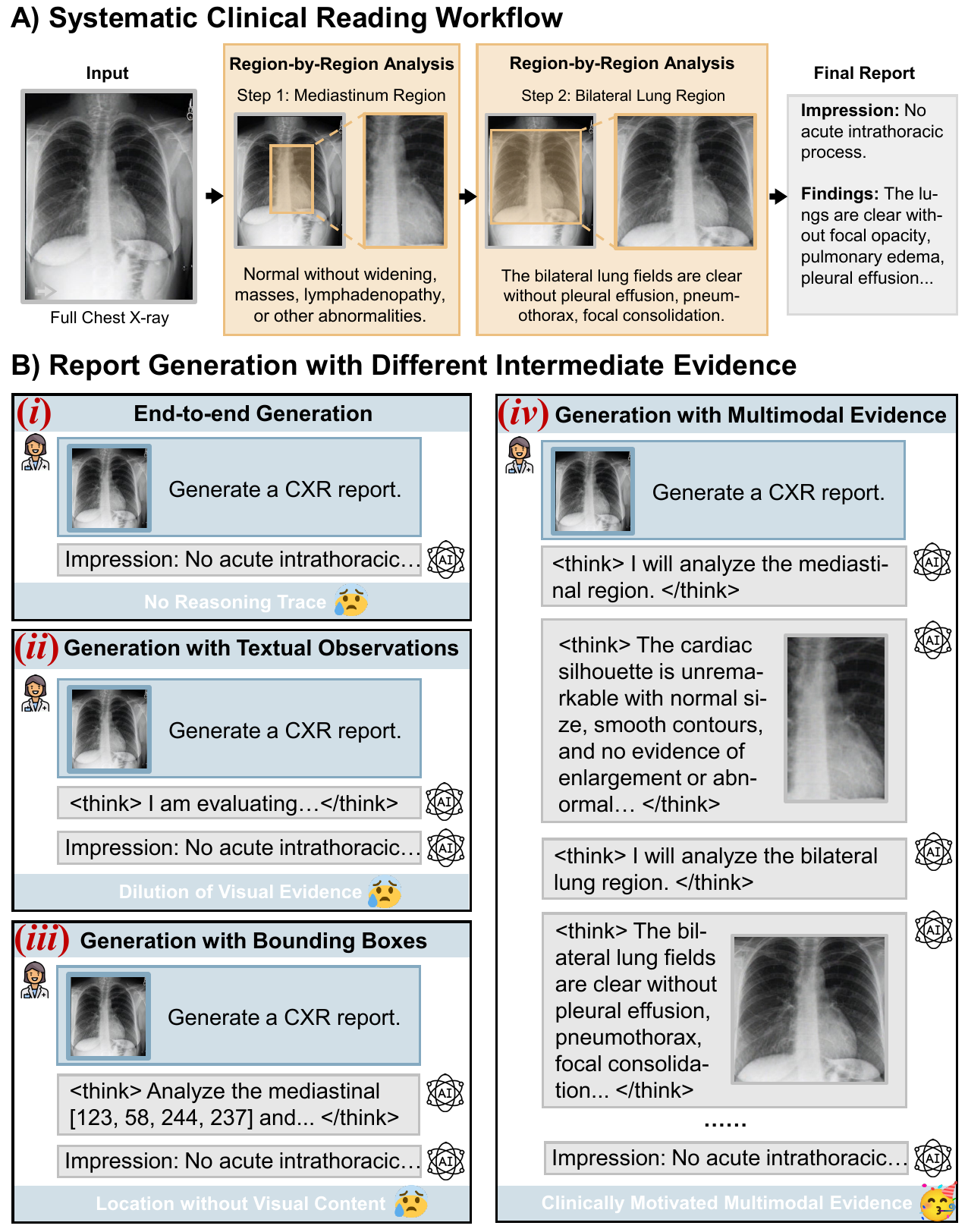}
    \caption{\textbf{Illustration of clinical reading and intermediate evidence for CXR report generation.} (A) Radiologists inspect a full chest X-ray through region-level analysis and synthesize these observations into a final report. (B) Different report generation settings use different forms of intermediate evidence: (i) end-to-end generation, (ii) textual observations, (iii) anatomical bounding boxes, and (iv) multimodal evidence.}
    \label{figworkflow}
\end{figure}

Chest X-ray (CXR) report generation aims to automatically produce clinically coherent radiology reports from medical images and has become a central task in medical vision-language modeling~\cite{jing2018automatic,li2018hybrid,miura2021improving}. Existing approaches have evolved from encoder-decoder frameworks such as R2Gen~\cite{chen2020generating} to recent large vision-language models (LVLMs)~\cite{wang2023r2gengpt,liu2025enhanced}. Despite substantial progress, most current paradigms still formulate report generation as a global image-to-text mapping problem, where a full CXR image is directly transformed into a report through autoregressive decoding (Fig.~\ref{figworkflow}, panel B(i)). Such supervision provides limited guidance on how localized findings are linked to the final report.

This limitation is especially important in chest X-ray interpretation, where reports are formed by integrating findings from multiple regions rather than from a single global summary. In clinical reading, radiologists inspect localized visual patterns, assess their clinical significance, and then synthesize these observations into the final report~\cite{north2024interpret}. As illustrated in Fig.~\ref{figworkflow}, panel A, regional observations are naturally associated with localized visual evidence during this process (\textit{\textcolor{orange}{orange box}}). Recent medical LVLMs have begun to explore richer forms of intermediate evidence. For example, LVMed-R2~\cite{wang2025lvmed} and MRG-R1~\cite{wang2025mrg} introduce textual observations to improve report generation quality and factual consistency (Fig.~\ref{figworkflow}, panel B(ii)), while models such as CheXagent~\cite{chen2024chexagent}, MAIRA-2~\cite{bannur2024maira}, and RadVLM~\cite{deperrois2025radvlm} incorporate anatomical grounding, regional cues, or broader medical assistance capabilities (Fig.~\ref{figworkflow}, panel B(iii)). However, textual observations provide limited visual grounding, and bounding-box coordinates indicate location without preserving localized image details. At the dataset level, existing radiology resources provide only fragmented supervision signals, such as report-level targets~\cite{johnson2019mimic}, anatomical grounding coordinates~\cite{liu2025gemex}, or grounded reasoning traces~\cite{liu2025gemexthink}, but rarely organize regional textual observations, localized image evidence, grounding annotations, and report targets into a unified multimodal report-generation-oriented structure. As a result, the utility of such structured multimodal evidence for chest X-ray report generation has not been systematically studied.

To better understand this issue, we first conduct a pilot comparison of different localized evidence formats for chest X-ray report generation, including textual observations, bounding boxes, localized image, and paired multimodal evidence. As shown in Fig.~\ref{figwmetric}, across multiple advanced closed-source LVLMs and evaluation metrics, the multimodal setting generally performs the best among the compared formats. This suggests that structured multimodal evidence is a promising supervision format for report generation, because combining regional textual observations with localized images can provide complementary support. However, existing radiology resources do not explicitly organize it in a reusable study-level form. Driven by these findings, we introduce \textbf{MMRad-22K}, a study-level chest X-ray report generation dataset organized around anatomy-guided multimodal evidence. Built upon MIMIC-CXR~\cite{johnson2019mimic} and GEMeX-ThinkVG~\cite{liu2025gemexthink}, MMRad-22K reorganizes fragmented supervision signals into report-oriented evidence units within each study, providing a structured multimodal resource for chest X-ray report generation. Finally, we evaluate MMRad-22K through adaptation experiments on a unified LVLM backbone Anole. The results show that using MMRad-22K improves report generation quality over standard end-to-end generation and textual evidence settings under the same framework. In addition, the adapted model achieves comparable performance relative to representative open-source LVLMs, despite not introducing a specialized model architecture.

Our contributions are summarized as follows. (1) We introduce \textbf{MMRad-22K}, a multimodal dataset specifically structured for CXR report generation, organizing regional textual observations, anatomical grounding coordinates, localized images, and report targets into a structured multimodal resource. (2) We present a task-oriented construction pipeline that reformulates fragmented localized supervision into report-generation-oriented evidence units, with multi-stage verification and clinician evaluation to improve clinical accuracy and consistency. (3) We provide empirical evidence that structured multimodal supervision is useful for CXR report generation: pilot comparisons suggest multimodal evidence is more effective than text-only or bounding box-based alternatives, and controlled adaptation experiments show that MMRad-22K supports stronger report generation than standard end-to-end and textual-evidence settings; under the same evaluation protocol, the adapted model also reaches a performance level comparable to several open-source LVLM references.

\section{Related Work}

\subsection{End-to-End CXR Report Generation}

Early methods mainly followed encoder-decoder image-to-text paradigms, including memory-driven frameworks such as R2Gen~\cite{chen2020generating}. More recently, LVLMs have been adapted to this task, as exemplified by SEI~\cite{liu2024structural}, FedMRG~\cite{metmer2025fedmrg}, R2GenGPT~\cite{wang2023r2gengpt}, and MLRG~\cite{liu2025enhanced}. Despite this progress, end-to-end image-to-report generation remains a standard formulation in chest X-ray report generation.

\subsection{Intermediate Evidence for CXR Report Generation}

\begin{figure*}[t]
    \centering
    \includegraphics[width=\textwidth]{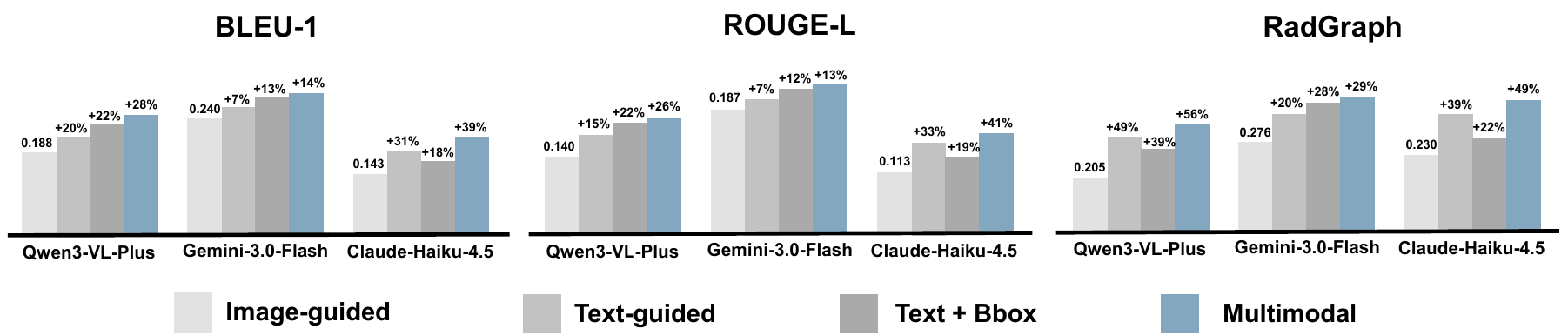}
    \caption{\textbf{Performance under different intermediate evidence formats.} Augmenting the standard full-image report generation setting with intermediate multimodal evidence generally yields the best or comparable performance across multiple LVLMs and evaluation metrics.}
    \label{figwmetric}
\end{figure*}

Recent medical LVLMs have begun to incorporate richer supervision signals beyond global image-report pairs. Some approaches improve reasoning and factuality through intermediate textual evidence, rationale-style supervision, or reflection strategies, including LVMed-R2~\cite{wang2025lvmed}, MRG-R1~\cite{wang2025mrg},and BoxMed-RL~\cite{jing2025reason}. Other frameworks emphasize grounding and region-aware supervision, such as CheXagent~\cite{chen2024chexagent}, MAIRA-2~\cite{bannur2024maira}, RadVLM~\cite{deperrois2025radvlm}, and ClinCoT~\cite{liu2026clincot}. These studies highlight the value of region-aware reasoning and grounding signals for medical vision-language modeling. However, such supervision is typically embedded in model architectures, training objectives, or multitask systems, rather than explicitly organized as a reusable report-generation-oriented resource that pairs textual observations with localized images.

\subsection{Radiology Datasets and Resources}

Several public datasets support medical vision-language learning. MIMIC-CXR~\cite{johnson2019mimic} mainly provides report-level supervision, while other resources introduce complementary signals such as anatomical grounding and reasoning traces~\cite{liu2025gemex,liu2025gemexthink,wu2021chest}. Although these datasets provide valuable supervision, their annotations remain fragmented across different tasks and resources. MMRad-22K is designed to bridge this gap by providing structured multimodal evidence for chest X-ray report generation.
%\section{What Form of Localized Evidence Best Supports Chest X-ray Report Generation?}

\section{Does Multimodal Evidence Help CXR Report Generation?}
\label{benefit}

Chest X-ray report generation requires models to transform visual observations into clinically coherent textual descriptions. In practice, intermediate supervision may be provided in different forms, such as localized image regions, textual observations, grounding coordinates, or combinations of these signals.

We conduct a pilot comparison of several intermediate evidence formats under the same report generation setting. For each study, the model always receives the full chest X-ray image, and we additionally provide evidence from the same anatomical region in one of the following formats: \textbf{(1) Image-guided:} a regional image crop; \textbf{(2) Text-guided:} a regional textual observation describing that region; \textbf{(3) Text + BBox:} the same regional textual observation augmented with anatomical grounding coordinates; and \textbf{(4) Multimodal:} both the regional textual observation and its corresponding image crop.

\noindent \textbf{Evaluation Setup.} We conduct the pilot study on 1, 097 CXR studies derived from MIMIC-CXR~\cite{johnson2019mimic} and GEMeX-ThinkVG~\cite{liu2025gemexthink}, where each study contains report targets together with region-level textual and visual evidence used to instantiate different evidence formats. We evaluate three advanced closed-source LVLMs, including Qwen3-VL-Plus~\cite{bai2025qwen3}, Gemini-3.0-Flash~\cite{GoogleGemini3Flash2025}, and Claude-Haiku-4.5~\cite{AnthropicClaude2025}, using NLG metrics BLEU~\cite{papineni2002bleu}, ROUGE-L~\cite{lin2004rouge}, and the clinical efficacy (CE) metric RadGraph~\cite{delbrouck2022improving}. Prompt templates and input examples are provided in Appendix~\ref{appbenefit}.

\begin{figure*}[t]
    \centering
    \includegraphics[width=\textwidth]{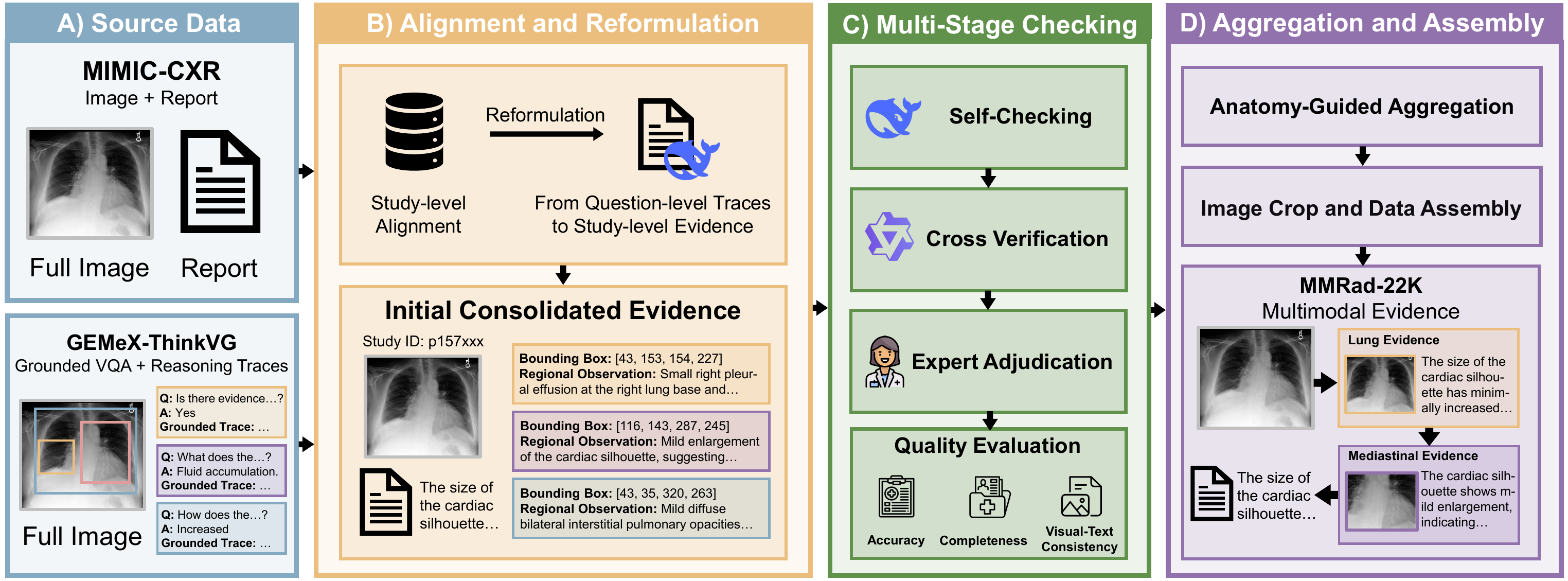}
    \caption{\textbf{Construction pipeline of MMRad-22K.} Starting from chest X-ray studies and grounded regional annotations, we reorganize fragmented question-level supervision into study-level report-oriented evidence units. The pipeline includes study-level alignment, evidence reformulation, multi-stage verification, quality assessment, and anatomy-guided aggregation. The final dataset unifies regional textual observations, anatomical coordinates, localized image evidence, and report targets into a structured multimodal resource for chest X-ray report generation.}
    \label{figpipeline}
\end{figure*}

\noindent \textbf{Multimodal localized evidence yields the most consistent gains.} Fig.~\ref{figwmetric} summarizes the performance of multiple advanced closed-source LVLMs under different localized evidence settings. Across models and metrics, multimodal format generally performs best or remains competitive in our pilot comparison. In contrast, adding anatomical bounding-box coordinates  yields smaller and less consistent gains over text-guided formats. These results suggest that localized textual and visual evidence can provide complementary support for CXR report generation.

However, existing radiology datasets provide only fragmented supervision signals, such as reports, grounding annotations, or textual rationales, rather than a unified report-generation-oriented multimodal evidence structure. To bridge this gap, we construct \textbf{MMRad-22K}, which organizes  textual observations, anatomical grounding coordinates, and localized image evidence within each study as structured multimodal supervision for CXR report generation.

\section{MMRad-22K Dataset}
\subsection{Data Construction}
\label{dataconstruction}

MMRad-22K is designed as a report-generation-oriented multimodal evidence dataset for chest X-ray reasoning. To construct this supervision, we leverage complementary signals from two study-aligned resources built upon the same radiographs, as shown in Fig.~\ref{figpipeline}, panel A. MIMIC-CXR~\cite{johnson2019mimic} provides chest X-ray images and report-level supervision, while GEMeX-ThinkVG~\cite{liu2025gemexthink} provides grounded VQA-style reasoning traces and anatomical bounding boxes. Our goal is to transform these fragmented grounded traces into anatomy-structured localized multimodal evidence suitable for study-level report generation.

\noindent \textbf{Study-Level Alignment and Evidence Reformulation.}
Grounded supervision in GEMeX-ThinkVG~\cite{liu2025gemexthink} is organized as independent question-specific traces, which are not directly suited for study-level report generation. As shown in Fig.~\ref{figpipeline}, panel B, we first align the grounded traces and anatomical bounding boxes from GEMeX-ThinkVG with their corresponding MIMIC-CXR~\cite{johnson2019mimic} studies, associating localized evidence with the full chest X-ray image and report target. GEMeX-ThinkVG contains 202,384 grounded VQA traces, which are reorganized after alignment into 21,994 study-level samples, with approximately 9 localized evidence traces per study on average. We then use DeepSeek-v3~\cite{liu2024deepseek} to reformulate these fragmented VQA traces into report-oriented evidence units, converting question-answer supervision into anatomy-associated regional observations for chest X-ray report generation.

\begin{figure*}[t]
    \centering
    \includegraphics[width=0.95\textwidth]{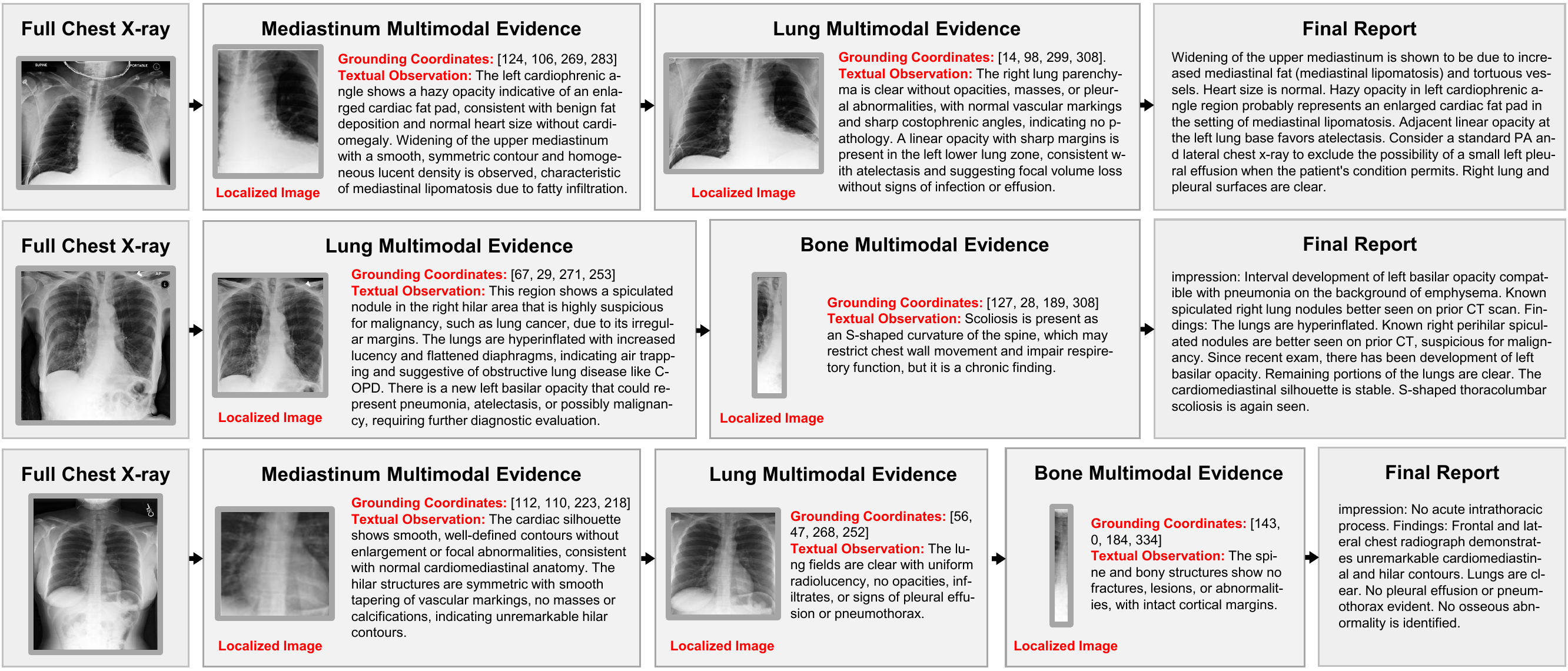}
    \caption{\textbf{Examples of MMRad-22K Dataset.} Each sample includes the full CXR image, structured multimodal evidence, and the final report.}
    \label{figdatacase}
\end{figure*}

\noindent \textbf{Clinical Consistency Preservation and Verification.}
During the reformulation process, LLMs may introduce summarization errors or hallucinated content. To improve consistency, we adopt a multi-stage verification pipeline, as shown in Fig.~\ref{figpipeline}, panel C. First, DeepSeek-v3~\cite{liu2024deepseek} performs a self-check by comparing the reformulated evidence units with the original grounded traces from the same study, mainly to remove unsupported content and improve local coherence. Second, we perform study-level consistency checking with Qwen2.5-72B~\cite{qwen2.5}, using the MIMIC-CXR reference report as a quality-control signal to identify potential contradictions between the reformulated evidence and the documented findings. The reference report is used only for verification rather than for generating new evidence content. Cases that remain uncertain after this step are further reviewed through clinician adjudication. A coarse summary of the verification pipeline is provided in Appendix~\ref{appveri}.

To assess the quality of the reformulated evidence units, we randomly sampled 900 cases (4\% of the full dataset) for blinded evaluation by two clinicians with radiology experience. Each case was independently rated on three 5-point Likert scales measuring \textbf{Clinical Accuracy}, \textbf{Completeness}, and \textbf{Visual-Text Consistency}. Across the sampled cases, the evidence units achieved average scores of 4.6/5.0, 4.3/5.0, and 4.9/5.0 for the three dimensions, respectively, with 96.7\%, 83.3\%, and 95.6\% of cases receiving scores of at least 4 from both clinicians. Inter-rater agreement measured by Cohen's $\kappa$ was 0.64, 0.50, and 0.56, respectively. The slightly lower completeness scores mainly reflect that some secondary findings described in the original reports are not explicitly covered by the available evidence units. Given the strong concentration of ratings in the high-score range, the percentage of cases jointly rated at least 4 by both clinicians provides a more direct indication of evidence quality than inter-rater agreement. Overall, the reformulated evidence units show good quality. Detailed evaluation rubrics are provided in Appendix~\ref{appscore}.

\begin{figure}[t]
    \centering
    \includegraphics[width=0.9\columnwidth]{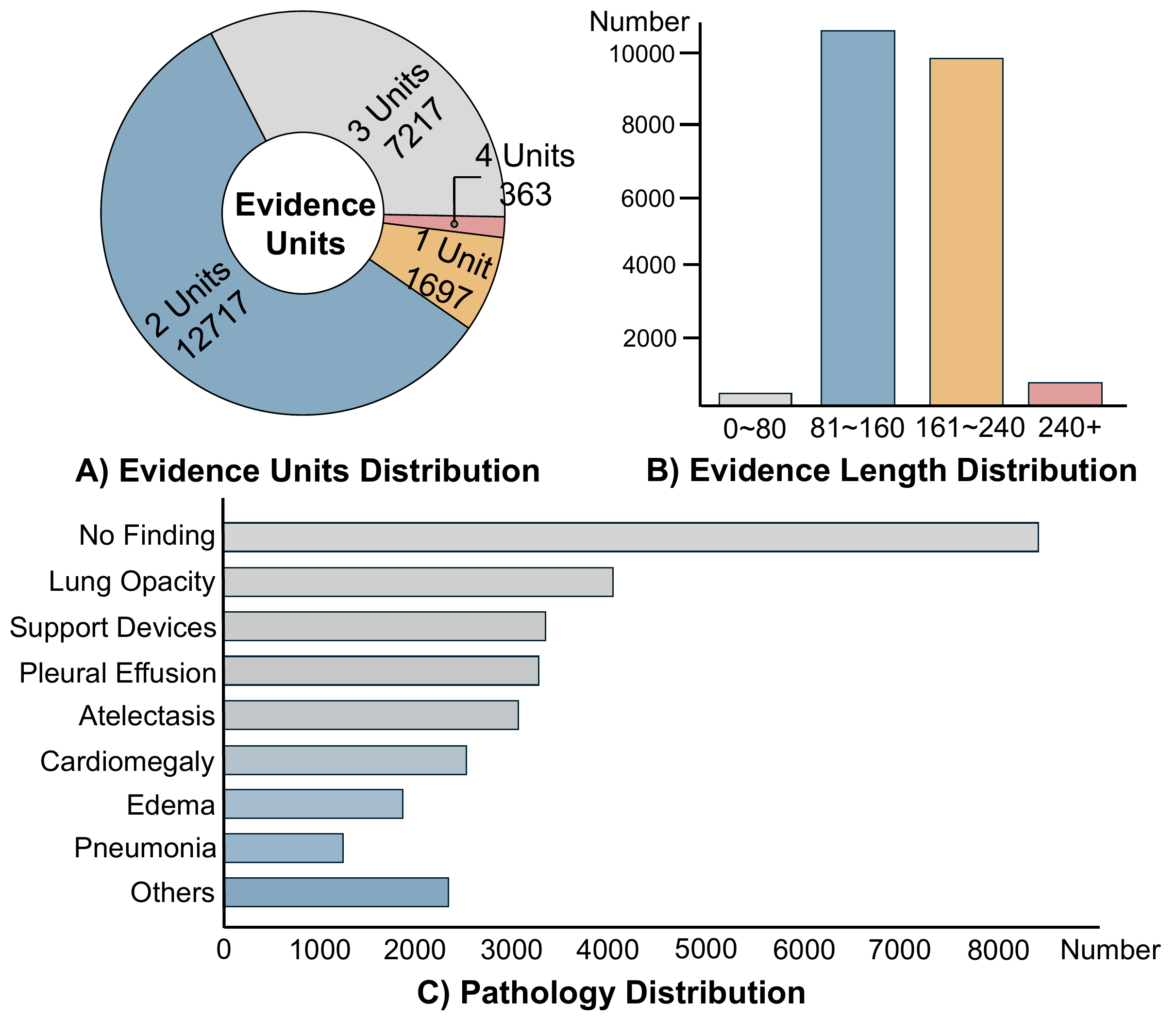}
    \caption{\textbf{Statistics of MMRad-22K.} The dataset contains structured regional evidence with diverse clinical findings and varying evidence lengths.}
    \label{figstatistics}
\end{figure}

\noindent \textbf{Anatomy-Guided Evidence Organization and Dataset Assembly.}
The reformulated evidence units are often fine-grained and heterogeneous in structure. To obtain a more consistent and learnable study-level representation, we organize the evidence into four coarse anatomy-guided groups: mediastinal, lung, bone, and other regions. Although this grouping is intentionally coarse, it provides a practical abstraction for model learning while remaining broadly consistent with classical chest X-ray interpretation frameworks in radiology textbooks and prior technical work~\cite{collins2012chest,wu2021chest}. Importantly, within each anatomical group, the original regional observations and their associated bounding boxes are retained and organized together rather than re-summarized. Based on the grouped bounding boxes, we further extract localized image crops from the original MIMIC-CXR image and pair them with the corresponding textual observations and coordinates to form multimodal evidence units for each region. Examples of MMRad-22K are shown in Fig.~\ref{figdatacase}, with additional examples provided in Appendix~\ref{moredataset}.

\begin{table}[t]
\centering
\caption{\textbf{Comparison with related chest X-ray datasets in terms of report targets and intermediate evidence.} MMRad-22K uniquely combines regional textual observations, anatomical grounding coordinates, structured multimodal evidence, and report-level targets within each study.}
\label{tabcomparison}
\fontsize{9}{11}\selectfont
\setlength{\tabcolsep}{2.5pt}
\resizebox{\columnwidth}{!}{
\begin{tabular}{lcccc}
\toprule
{Dataset} 
& \makecell[c]{{Report}\\{Target}}
& \makecell[c]{{Textual}\\{Evidence}}
& \makecell[c]{{Anatomical}\\{Grounding}}
& \makecell[c]{{Multimodal}\\{Evidence}} \\
\midrule
EHRXQA
& \xmark & \xmark & \xmark & \xmark \\
MIMIC-CXR
& \cmark & \xmark & \xmark & \xmark \\
GEMeX
& \xmark & \cmark & \cmark & \xmark \\
Chest ImaGenome 
& \xmark & \cmark & \cmark & \xmark \\
GEMeX-ThinkVG
& \xmark & \cmark & \cmark & \xmark \\
\midrule
\rowcolor[gray]{0.95}
\textbf{MMRad-22K} 
& \cmark & \cmark & \cmark & \cmark \\
\bottomrule
\end{tabular}
}
\end{table}

\subsection{Data Statistics}

As shown in Fig.~\ref{figstatistics}, MMRad-22K contains 21,994 study-level multimodal evidence samples from 7,975 patients. Most studies contain 2–3 regional evidence groups (Fig.~\ref{figstatistics}, panel A). The evidence text is also relatively rich, with most samples ranging from 81 to 240 words in length (Fig.~\ref{figstatistics}, panel B). The dataset covers a broad range of findings based on CheXpert~\cite{irvin2019chexpert} labels (Fig.~\ref{figstatistics}, panel C), including lung opacity, pleural effusion, atelectasis, cardiomegaly, edema, and pneumonia, in addition to normal studies (No Finding). Additional anatomy-group statistics are provided in the Appendix~\ref{dataset}.

\subsubsection{Data Comparison}

As shown in Table~\ref{tabcomparison}, existing radiology datasets provide only partial supervision for chest X-ray report generation. MIMIC-CXR~\cite{johnson2019mimic} and EHRXQA~\cite{bae2023ehrxqa} provide report-level targets and multimodal QA pairs, respectively, but do not include explicit intermediate evidence. Chest ImaGenome~\cite{wu2021chest} provides bounding boxes and localized labels, but does not organize them into coherent study-level multimodal evidence for report composition. GEMeX~\cite{liu2025gemex} and GEMeX-ThinkVG~\cite{liu2025gemexthink} provide grounded question-level textual supervision, yet their annotations remain organized as VQA-oriented explanations or reasoning traces rather than report-generation-oriented regional evidence.

In contrast, MMRad-22K is not a simple concatenation of existing datasets, but a task-oriented restructuring of fragmented supervision into a unified multimodal resource for report generation. Specifically, it converts question-level grounded traces into study-level, structured evidence units that are aligned with the original report target. Each sample is organized around a single study and explicitly links regional textual observations, anatomical coordinates, localized images, and the final report within the same report-generation context, which is not directly supported by prior datasets.

\section{MMRad-22K for CXR Report Generation}

Section~\ref{benefit} suggests that localized multimodal evidence benefits CXR report generation. We next evaluate whether MMRad-22K effectively supports LVLM adaptation and provides advantages beyond standard report supervision.

\subsection{Training Settings}

\noindent \textbf{Dataset Split.}
We randomly split MMRad-22K at the patient level. Following this protocol, 237 patients are reserved for testing, while the remaining 7738 patients are used for training.

\begin{table*}[t]
\centering
\caption{\textbf{Chest X-ray report generation results under different evidence settings on the Anole backbone.} \textbf{Bold} denotes the best result. Abbreviations: B-n: BLEU-n; MTR: METEOR; R-L: ROUGE-L; RG: RadGraph. }
\label{ablation}
\resizebox{\textwidth}{!}{%
\begin{tabular}{l ccccc ccccc}
\toprule
\multirow{2}{*}{{Setting}} & \multicolumn{5}{c}{{NLG Metrics}} & \multicolumn{5}{c}{{CE Metrics}} \\
\cmidrule(lr){2-6} \cmidrule(lr){7-11}
 & {B-1$\uparrow$} & {B-2$\uparrow$} & {B-3$\uparrow$} & {MTR$\uparrow$} & {R-L$\uparrow$} & {RadCliQ$\downarrow$} & {RG$_e$$\uparrow$} & {RG$_{er}$$\uparrow$} & {RG$_{ber}$$\uparrow$} & {RaTE$\uparrow$} \\
\midrule
Zero-shot & 0.096 & 0.034 & 0.007 & 0.104 & 0.095 & 2.102 & 0.082 & 0.070 & 0.041 & 0.464 \\
End-to-end & 0.126 & 0.040 & 0.008 & 0.107 & 0.103 & 1.821 & 0.112 & 0.099 & 0.067 & 0.462 \\
Textual & 0.175 & 0.075 & 0.032 & 0.144 & 0.150 & 1.541 & 0.149 & 0.133 & 0.092 & 0.493 \\
\rowcolor[gray]{0.95} Grounded & 0.192 & \textbf{0.091} & \textbf{0.046} & 0.166 & \textbf{0.173} & 1.409 & 0.180 & 0.161 & 0.112 & 0.528 \\
\rowcolor[gray]{0.95} Generated & \textbf{0.193} & 0.090 & 0.043 & \textbf{0.167} & 0.172 & \textbf{1.391} & \textbf{0.187} & \textbf{0.173} & \textbf{0.123} & \textbf{0.529} \\
\bottomrule
\end{tabular}%
}
\end{table*}

\begin{figure}[t]
    \centering
    \includegraphics[width=0.45\textwidth]{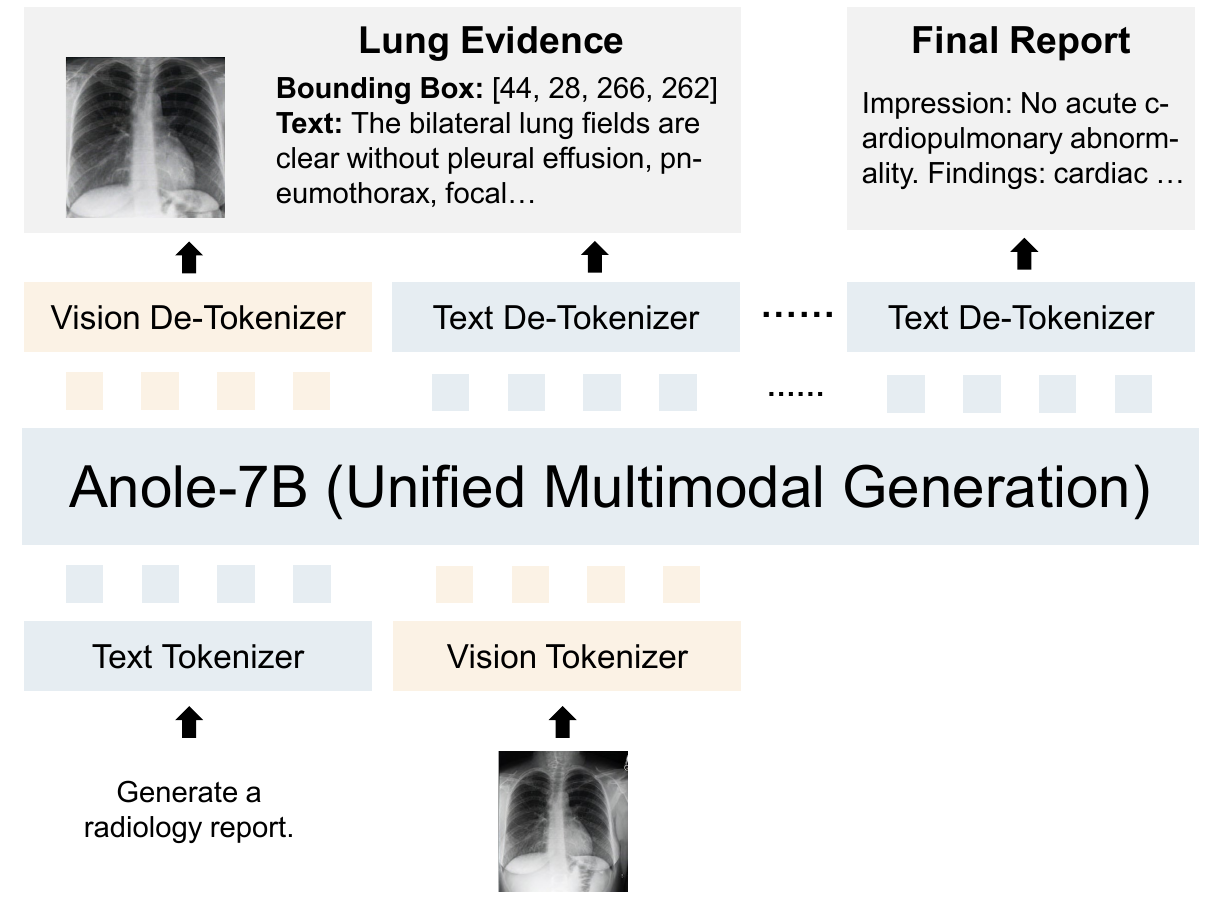}
    \caption{\textbf{Unified multimodal framework based on MMRad-22K.} Each training sample is organized as a unified multimodal sequence.}
    \label{figanole}
\end{figure}

\noindent \textbf{Baseline Model.} We adopt Anole-7B~\cite{chern2024anole} as the base unified LVLM because of its native multimodal autoregressive generation capability, as shown in Fig.~\ref{figanole}. Unlike conventional LVLMs designed primarily for visual understanding, Anole enables unified generation over both textual and visual tokens within a single autoregressive framework, making it suitable for modeling the multimodal evidence structures in MMRad-22K and the final radiology report within one generation process. We fine-tune the model using LoRA~\cite{hu2022lora} applied to the query, key, and value projections of the attention modules with rank $r=16$ and scaling factor $\alpha=32$, leveraging the training framework established in~\cite{chern2025thinking}. Training is conducted for 50,000 steps on two NVIDIA H800 GPUs using a learning rate of $1\times10^{-5}$ and batch size 2.

%\noindent \textbf{Training Framework.}
%Figure~\ref{figanole} illustrates the overall training framework. Each training sample consists of a full chest X-ray image, multiple evidence units, and the corresponding radiology report. The model is trained in a unified autoregressive manner to model these multimodal evidence structures together with the final report target.

\noindent \textbf{Evaluation Metrics.}
We evaluate report generation quality using standard NLG metrics, including BLEU~\cite{papineni2002bleu}, METEOR~\cite{banerjee2005meteor}, and ROUGE-L~\cite{lin2004rouge}, together with clinical efficacy (CE) metrics including RadGraph~\cite{delbrouck2022improving}, RadCliQ~\cite{yu2023evaluating}, and RaTEScore~\cite{zhao2024ratescore}. Detailed descriptions of the evaluation metrics are provided in the Appendix~\ref{evaluation_metrics}.

\subsection{Controlled Comparison of Supervision Settings on the Anole Backbone}

Starting from the original zero-shot Anole baseline, we compare five settings while keeping the report generation task unchanged. \textbf{Zero-shot} directly generates a report without task-specific adaptation. \textbf{End-to-end} adapts the model to generate the report directly from the full chest X-ray image. \textbf{Textual} first generates regional textual evidence and then produces the final report. \textbf{Generated} further generates localized visual evidence in Anole's native autoregressive generation space together with regional textual evidence before report generation. \textbf{Grounded} first generates regional textual evidence and predicts anatomical regions, then maps the predicted regions back to the source chest X-ray and extracts localized image crops as grounded visual evidence before report generation. Notably, the intermediate evidence is generated or predicted by the model itself rather than provided as external prompts. Additional details on evidence construction and the difference between Generated and Grounded are provided in Appendix~\ref{evidence_settings}.

\begin{table*}[t]
\centering
\caption{\textbf{External references for contextualizing the effectiveness of MMRad-22K.} All the models are evaluated under the same chest X-ray report generation task, test split, and evaluation protocol, while Anole-MMRad is adapted using MMRad-22K. Anole-MMRad reaches a performance level on par with representative open-source LVLM references, further supporting the effectiveness of MMRad-22K for adapting LVLMs to chest X-ray report generation. \textbf{Bold} denotes the best result. Abbreviations: Param.: activated parameters; G: general; M: medical; B-n: BLEU-n; MTR: METEOR; R-L: ROUGE-L; RG: RadGraph. }
\label{open_comparison}
\resizebox{\textwidth}{!}{%
\begin{tabular}{l cc cccc cccc}
\toprule
\multirow{2}{*}{{Models}} & \multirow{2}{*}{{Param.}} & \multirow{2}{*}{{Domain}} & \multicolumn{4}{c}{{NLG Metrics}} & \multicolumn{4}{c}{{CE Metrics}} \\
\cmidrule(lr){4-7} \cmidrule(lr){8-11}
 &  & & {B-1$\uparrow$} & {B-2$\uparrow$} & {MTR$\uparrow$} & {R-L$\uparrow$} & {RadCliQ$\downarrow$} & {RG$_e$$\uparrow$} & {RG$_{er}$$\uparrow$} & {RaTE$\uparrow$} \\
\midrule
\multicolumn{11}{l}{\textcolor{gray!80}{\textit{Understanding-Only Large Language Models}}} \\
Deepseek-VL2 & 4.5B & G & 0.105 & 0.033 & 0.090 & 0.103 & 1.888 & 0.084 & 0.075 & 0.495 \\
Qwen3-VL & 8B & G & 0.167 & 0.054 & 0.125 & 0.133 & 1.357 & 0.158 & 0.144 & 0.530 \\
InternVL3.5 & 8B & G & 0.179 & 0.059 & 0.135 & 0.147 & \textbf{1.316} & 0.152 & 0.145 & \textbf{0.532} \\
LLaVA-Med-1.5 & 7B & M & 0.127 & 0.011 & 0.076 & 0.100 & 1.755 & 0.072 & 0.063 & 0.426 \\
\midrule
\multicolumn{11}{l}{\textcolor{gray!80}{\textit{Unified Large Language Models}}} \\
Bagel & 7B & G & 0.111 & 0.040 & 0.131 & 0.113 & 1.602 & 0.157 & 0.135 & 0.525 \\
UniMedVL & 14B & M & 0.157 & 0.054 & 0.147 & 0.128 & 1.504 & 0.157 & 0.145 & 0.519 \\
\rowcolor[gray]{0.95} Anole & 7B & G & 0.096 & 0.034 & 0.104 & 0.095 & 2.102 & 0.082 & 0.070 & 0.464 \\
\rowcolor[gray]{0.95} \textbf{Anole-MMRad} & 7B & M & \textbf{0.193} & \textbf{0.090} & \textbf{0.167} & \textbf{0.172} & 1.391 & \textbf{0.187} & \textbf{0.173} & 0.529 \\
\bottomrule
\end{tabular}%
}
\end{table*}

The \textit{Zero-shot} Anole baseline performs poorly across evaluation metrics, indicating that the original general-domain unified model is not directly suited to chest X-ray report generation. Adapting the model in the \textit{End-to-end} setting already yields clear improvements, showing that task-specific adaptation is important in this domain.

Modeling regional textual evidence in the \textit{Textual} setting further improves performance over the end-to-end baseline, suggesting that intermediate textual evidence provides useful structured support beyond direct report generation from the full image alone. Extending this intermediate evidence to multimodal evidence yields additional gains. Both Multimodal settings \textit{Generated} and \textit{Grounded} outperform the text-only setting, further supporting the value of combining textual evidence with localized visual evidence, indicating that multimodal evidence derived from MMRad-22K provides complementary support for report generation.

Overall, these results suggest that MMRad-22K provides useful structured supervision for chest X-ray report generation. Relative to standard end-to-end adaptation, textual evidence improves report generation quality, and localized visual evidence brings further complementary gains.

\subsection{MMRad-22K Enables Effective Adaptation to CXR Report Generation}

To provide external context for the effectiveness of MMRad-22K, we additionally report the performance of the adapted Anole model (Anole-MMRad) together with several representative open-source LVLM references, including general-domain models DeepSeek-VL2~\cite{wu2024deepseek}, Qwen3-VL~\cite{bai2025qwen3}, InternVL3.5~\cite{wang2025internvl35}, Anole~\cite{chern2024anole}, and Bagel~\cite{deng2025emerging}, as well as medical-domain models LLaVA-Med-1.5~\cite{li2023llavamed} and UniMedVL~\cite{ning2025unimedvl}. Some of these reference models have prior exposure to medical report-generation-related training during their original development.

Table~\ref{open_comparison} shows that adapting Anole with MMRad-22K yields substantial gains over its original zero-shot performance. Under the same evaluation protocol, Anole-MMRad also reaches a performance level on par with representative open-source LVLM references. These results further support the practical utility of MMRad-22K as a multimodal supervision resource for adapting LVLMs to chest X-ray report generation.

\subsection{Case Study}

\begin{figure}[t]
    \centering
    \includegraphics[width=0.43\textwidth]{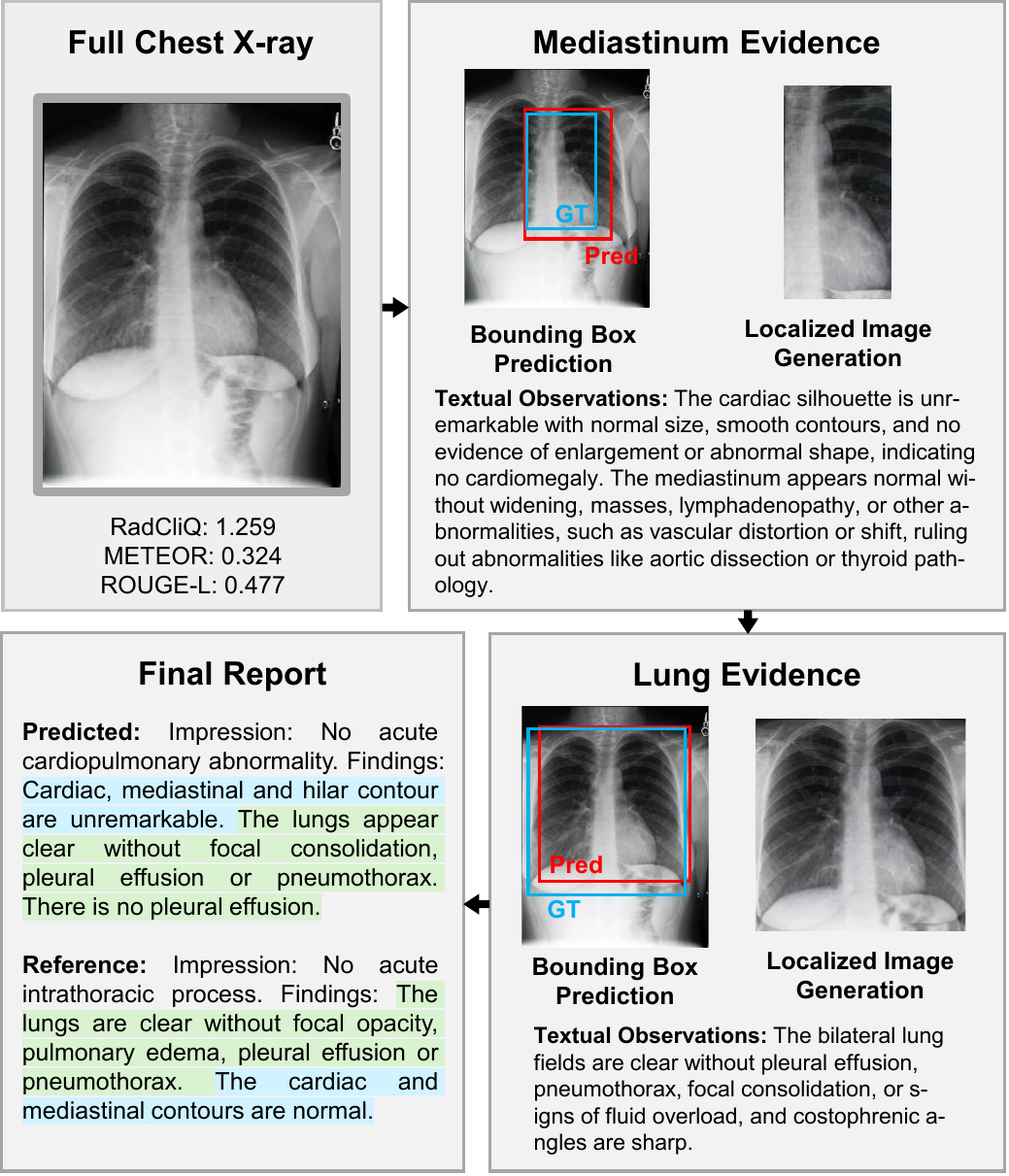}
    \caption{\textbf{Qualitative analysis by Anole-MMRad}. Green shading and blue shading represent consistent descriptions for the lung and mediastinal region.}
    \label{generatedstudy}
\end{figure}

Figure~\ref{generatedstudy} illustrates the model's systematic diagnostic trace on a representative CXR. The model first examines the mediastinal region by generating a localized image and textual observations that rule out cardiac and mediastinal abnormalities. It then transitions to the lung region where it identifies clear lung fields with high spatial accuracy. The final report closely aligns with the reference report. Additional case studies and examples of failure cases are provided in the Appendix~\ref{cases}.

\section{Conclusion}

In this work, we introduce MMRad-22K, a study-level structured multimodal dataset that unifies regional textual observations, anatomical grounding coordinates, localized images, and report targets into a structured format. Experiments show that this structured supervision is useful for CXR report generation: multimodal evidence outperforms weaker evidence formats in pilot comparisons, and adapting a unified LVLM with MMRad-22K improves over standard end-to-end and textual-evidence settings while reaching a performance level comparable to several open-source LVLM references. Overall, MMRad-22K provides a practical structured multimodal resource for training and evaluating CXR report generation models.

\section*{Limitations}

Despite the encouraging results provided by MMRad-22K, this study has several limitations that open avenues for future work. First, the dataset is constructed by reorganizing existing report-aligned public resources rather than through new prospective expert annotation collected specifically for study-level report generation. Although we apply multi-stage verification and clinician evaluation, the resulting evidence units may still reflect source reporting conventions and annotation biases. Second, MMRad-22K has imbalanced coverage across anatomical regions and findings, with lung- and mediastinal-related evidence accounting for the majority of samples, which may limit representation of rarer regions or subtle abnormalities. Third, because the dataset is derived from MIMIC-CXR, it remains to be validated whether the observed gains generalize to other institutions, reporting styles, and real clinical workflows.

\section*{Ethical Considerations}

MMRad-22K is intended solely as a research resource for developing and evaluating chest X-ray report generation systems. It is derived from existing public radiology resources and should not be used as a clinical decision-making tool or as a substitute for expert interpretation. As with other medical vision-language resources, potential risks include propagation of source reporting biases, incomplete coverage of rare findings, and over-reliance on automatically generated outputs in safety-critical settings. We therefore emphasize that models trained on MMRad-22K are for research use only, and their clinical reliability and generalization to new institutions remain to be established.

{
\bibliography{acl_latex}
}

\appendix
\clearpage
\section{Appendix}

\subsection{More Information about Pilot Study}
\label{appbenefit}

\subsubsection{Prompt Templates}

We use a shared report-generation objective across all settings, with common output requirements and the same report format. Because the additional evidence differs in modality, the prompt wording is minimally adapted to describe the corresponding input format  (\textit{\textcolor{red}{red mark}}). Thus, the comparison keeps the task objective and reporting constraints fixed while varying the form of additional region-level evidence.

\begin{figure}[h]
    \centering
    \includegraphics[width=\columnwidth]{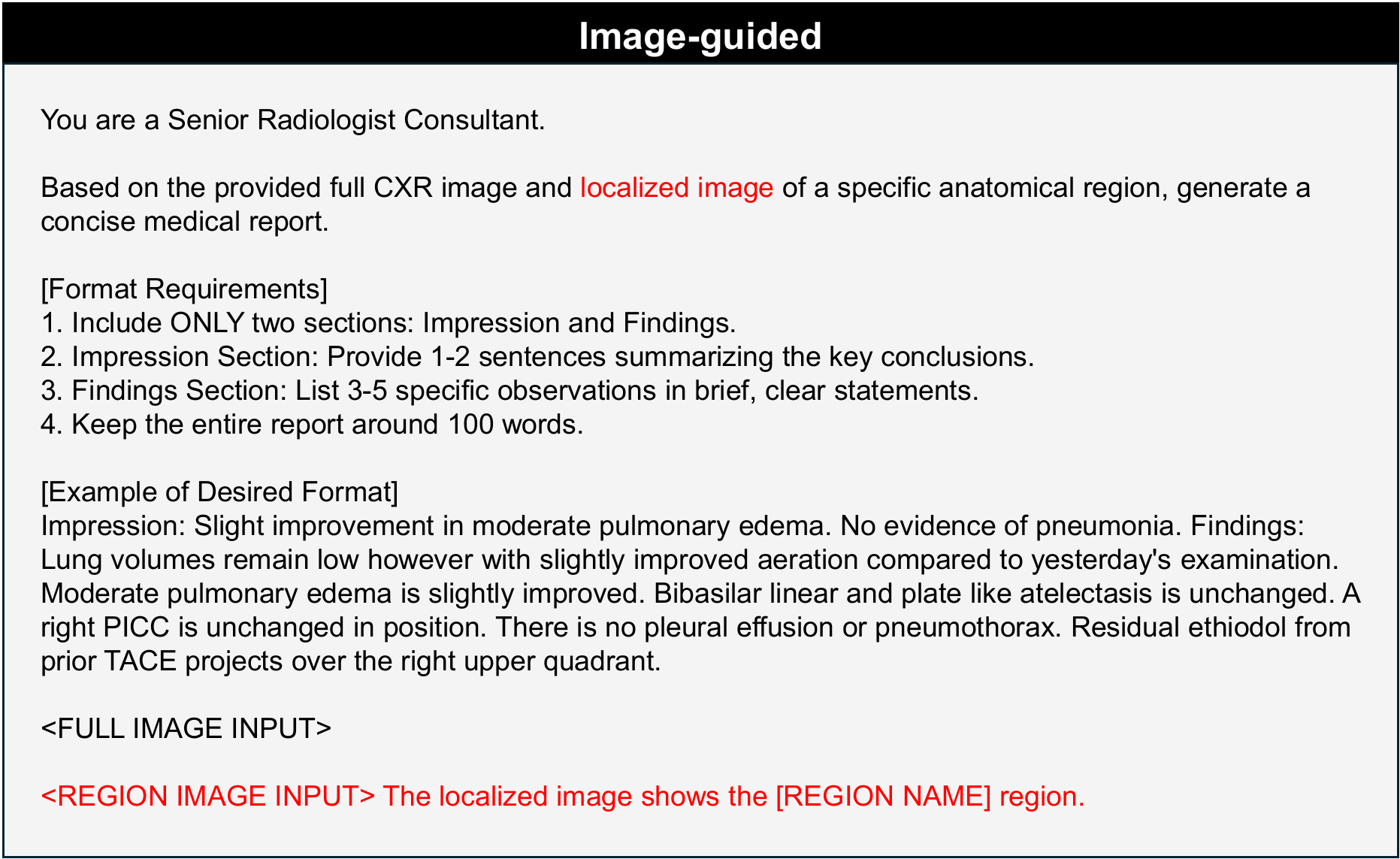}
\end{figure}

\begin{figure}[h]
    \centering
    \includegraphics[width=\columnwidth]{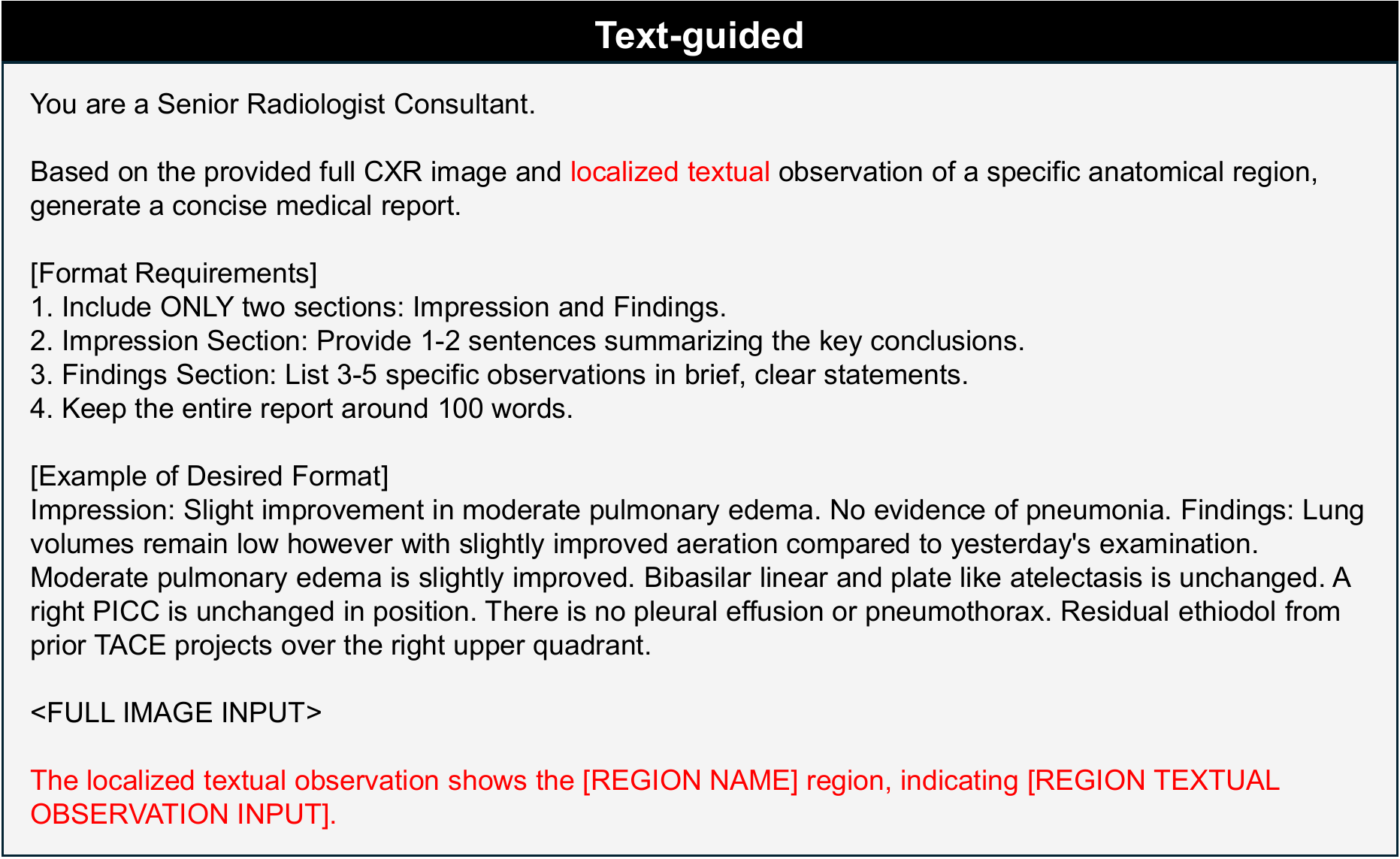}
\end{figure}

\begin{figure}[h]
    \centering
    \includegraphics[width=\columnwidth]{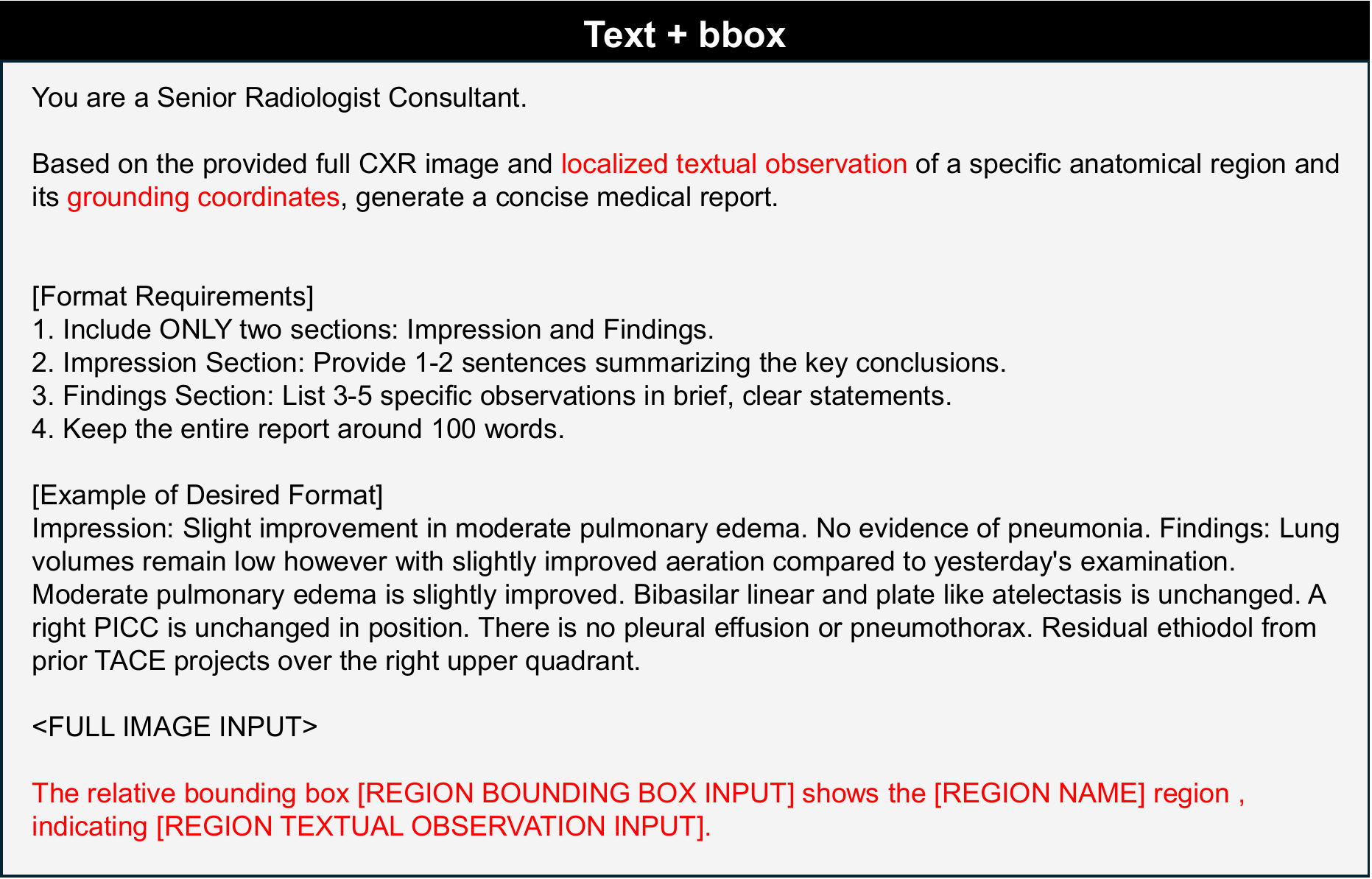}
\end{figure}

\begin{figure}[h]
    \centering
    \includegraphics[width=\columnwidth]{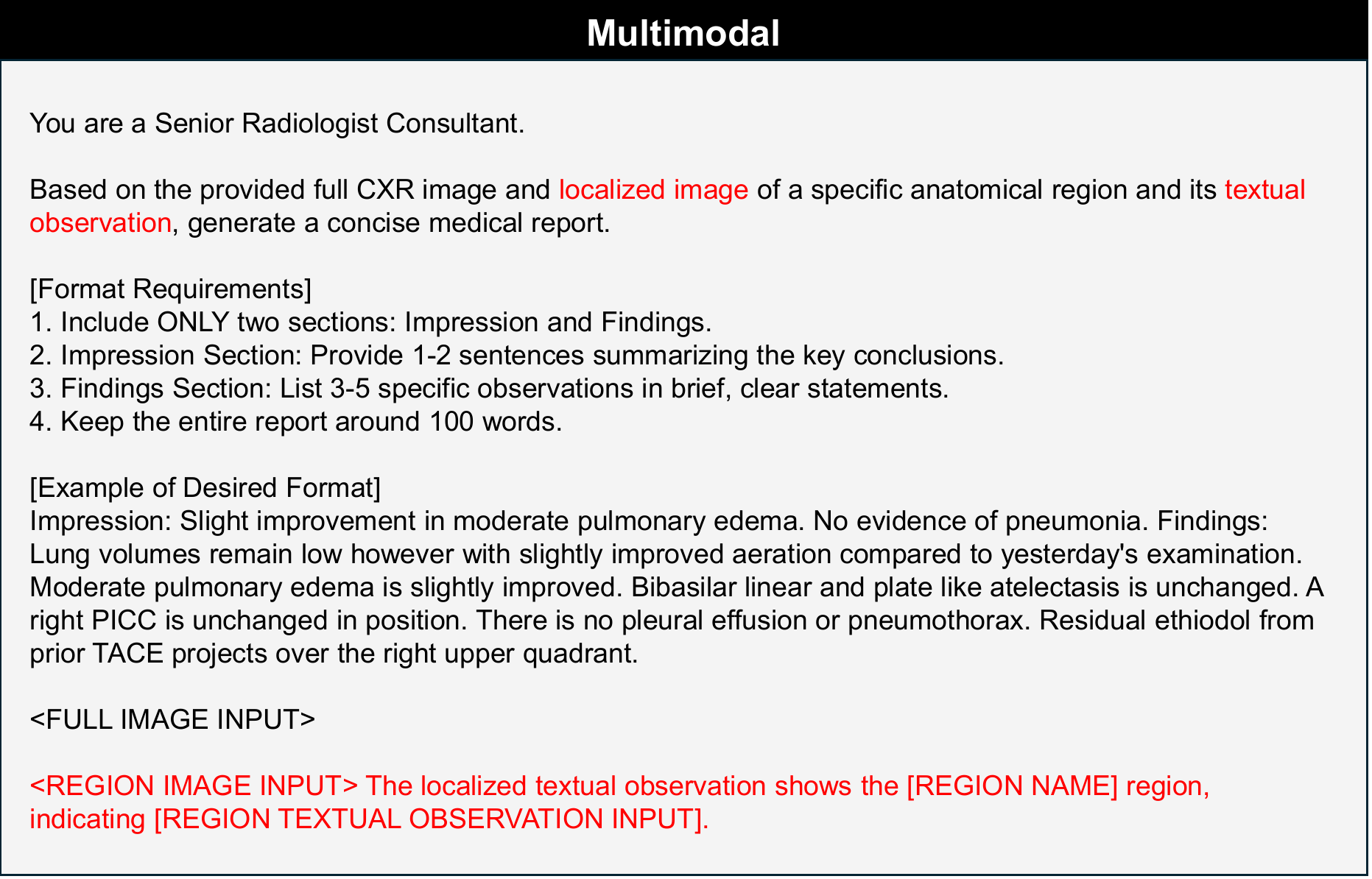}
\end{figure}

\subsubsection{Input Example}

We provide one example to illustrate how the same study case is instantiated under different evidence formats. The example includes the full chest X-ray image together with its region-level evidence, showing how image-guided, text-guided, text + bbox, and multimodal inputs differ while sharing the same report-generation target.

\begin{figure}[h]
    \centering
    \includegraphics[width=\columnwidth]{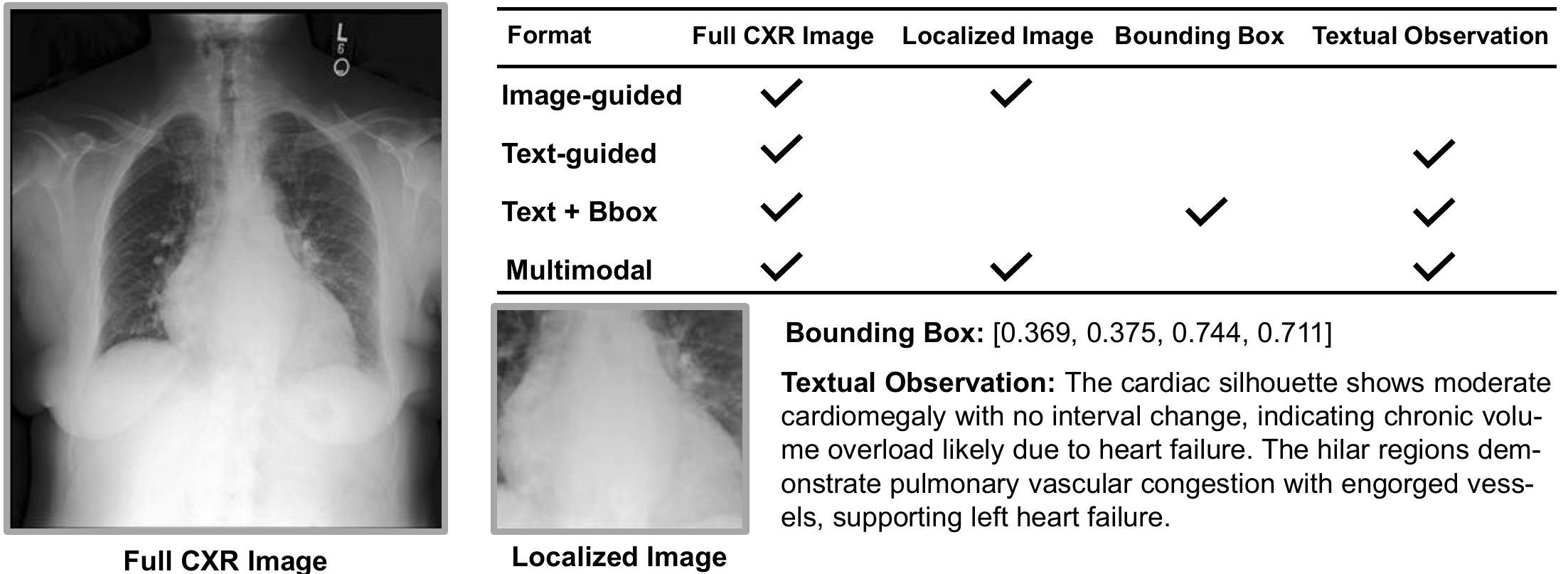}
\end{figure}

\subsection{More Information about Dataset Construction}
\label{appconstruction}

\subsubsection{Summary of the Verification Pipeline}
\label{appveri}
Approximately 14\% of reformulated evidence units were revised during the self-check stage. A coarse error breakdown suggests that about 9\% involved omission of source-supported content, around 4\% involved hallucinated or unsupported statements, and roughly 1\% involved mismatches between bounding boxes and the corresponding textual observations. In the cross-check stage, Qwen2.5-72B was used as a verifier to compare the reformulated evidence against the corresponding study-level report. About 2\% of cases were flagged at this stage for potential report-level inconsistencies or uncertain anatomical correspondence and were sent to clinician adjudication.

\subsubsection{Clinician Evaluation Rubrics}
\label{appscore}

\textbf{Clinical Accuracy} evaluates whether the regional observations are medically correct and consistent with the reference study findings, as shown in Table~\ref{tab:clinical_accuracy_rubric}. \textbf{Completeness} measures whether the reformulated evidence units cover the clinically important findings documented in the study, as shown in Table~\ref{tab:completeness_rubric}. \textbf{Visual-Text Consistency} evaluates whether each regional text description is supported by, and spatially aligned with, its associated localized image crop, as shown in Table~\ref{tab:visual_text_rubric}.

\begin{table}[h]
\centering
\caption{Rubric for clinical accuracy evaluation.}
\label{tab:clinical_accuracy_rubric}
\begin{tabularx}{\columnwidth}{c X}
\toprule
\textbf{Score} & \textbf{Clinical Accuracy} \\
\midrule
5 & Fully clinically accurate and well aligned with the reference report. \\
4 & Clinically accurate with only negligible imprecision. \\
3 & Generally correct but contains minor inaccuracies or ambiguous descriptions. \\
2 & Multiple clinically inaccurate or misleading observations. \\
1 & Major medical errors or hallucinated findings inconsistent with the reference report. \\
\bottomrule
\end{tabularx}
\end{table}

\begin{table}[h]
\centering
\caption{Completeness rubric.}
\label{tab:completeness_rubric}
\begin{tabularx}{\columnwidth}{c X}
\toprule
\textbf{Score} & \textbf{Completeness} \\
\midrule
5 & Comprehensive coverage of clinically relevant findings described in the reference report. \\
4 & Largely complete with only minor omissions. \\
3 & Covers major findings but misses some secondary observations. \\
2 & Several clinically relevant findings are omitted. \\
1 & Most clinically important findings are missing. \\
\bottomrule
\end{tabularx}
\end{table}

\begin{table}[h]
\centering
\caption{Rubric for visual-text consistency evaluation.}
\label{tab:visual_text_rubric}
\begin{tabularx}{\columnwidth}{c X}
\toprule
\textbf{Score} & \textbf{Visual-Text Consistency} \\
\midrule
5 & Strong visual-text correspondence with clear and well-localized visual support. \\
4 & Good visual-text alignment with only minor mismatch or ambiguity. \\
3 & Partial alignment between the localized image evidence and the textual observation. \\
2 & Weak or unclear correspondence between the image evidence and the text. \\
1 & The localized image evidence is inconsistent with the corresponding textual observation. \\
\bottomrule
\end{tabularx}
\end{table}

\subsection{More Information about the MMRad-22K Dataset}
\label{dataset}

\subsubsection{Additional Dataset Examples}
\label{moredataset}

\begin{figure*}[t]
    \centering
    \caption{\textbf{More Examples of MMRad-22K Dataset.} Each sample includes the full CXR image, structured multimodal evidence, and the final report.}
    \includegraphics[width=\textwidth]{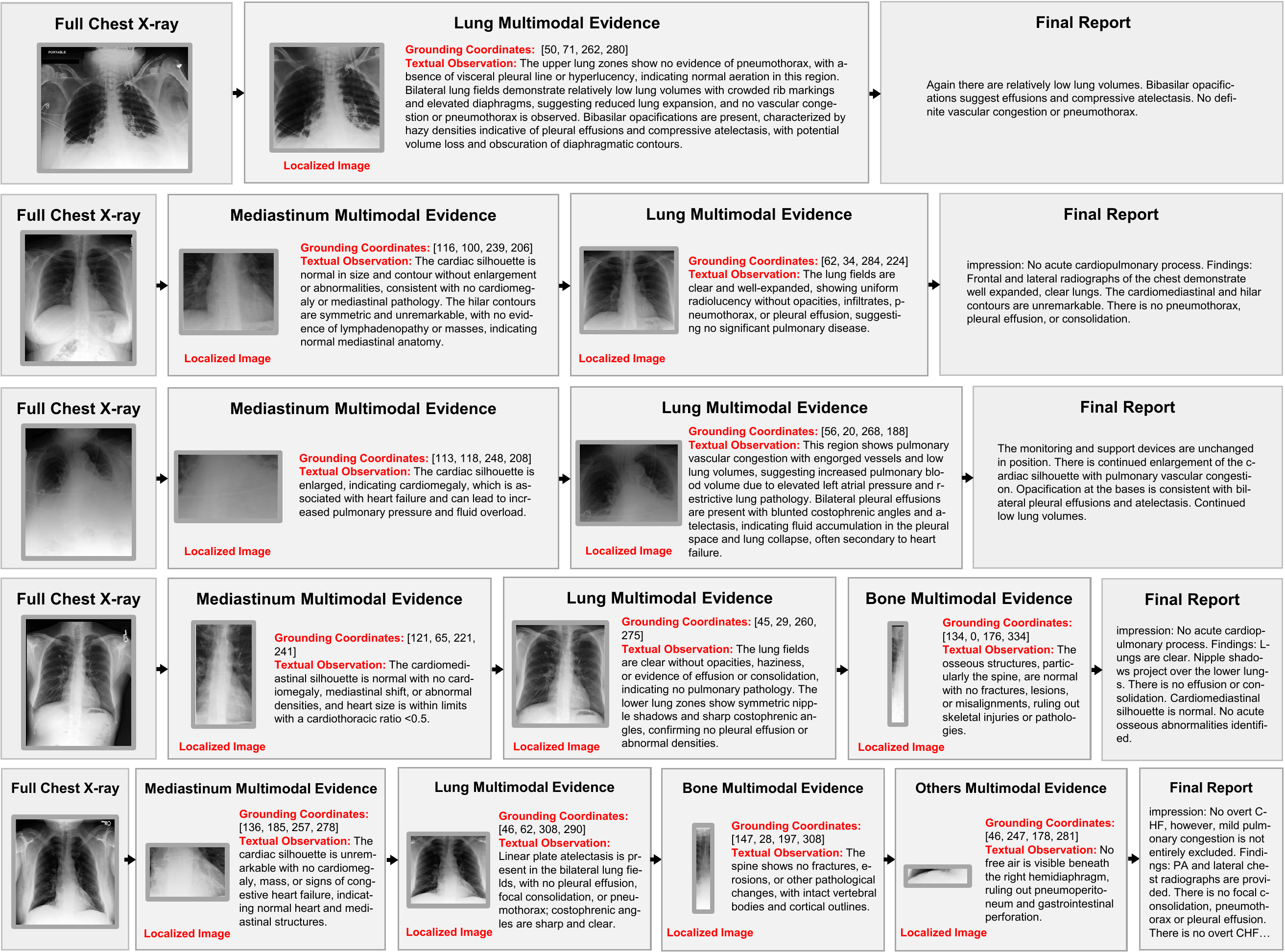}
    \label{moresamples}
\end{figure*}

Fig.~\ref{moresamples} presents additional examples of MMRad-22K. Each sample includes the full CXR image, one to four anatomy-guided multimodal evidence units, and the corresponding final report. These examples illustrate the study-level organization of the dataset, where localized image crops, textual observations, and grounding coordinates are grouped into clinically relevant regions and linked to the report target. They also show the variability of MMRad-22K across studies, ranging from simpler cases with a single evidence unit to more complex cases involving multiple anatomical groups such as lung, mediastinal, bone, and other regions.

\subsubsection{Anatomy-group Statistics}

\begin{figure*}[t]
    \centering
    \caption{\textbf{Fine-grained anatomical region distribution in MMRad-22K.} For readability, only regions with frequency greater than 1\% are shown. The distribution is long-tailed, with lung- and mediastinal-related regions predominating, while bone, abdominal, and support-device regions remain represented.}
    \includegraphics[width=\textwidth]{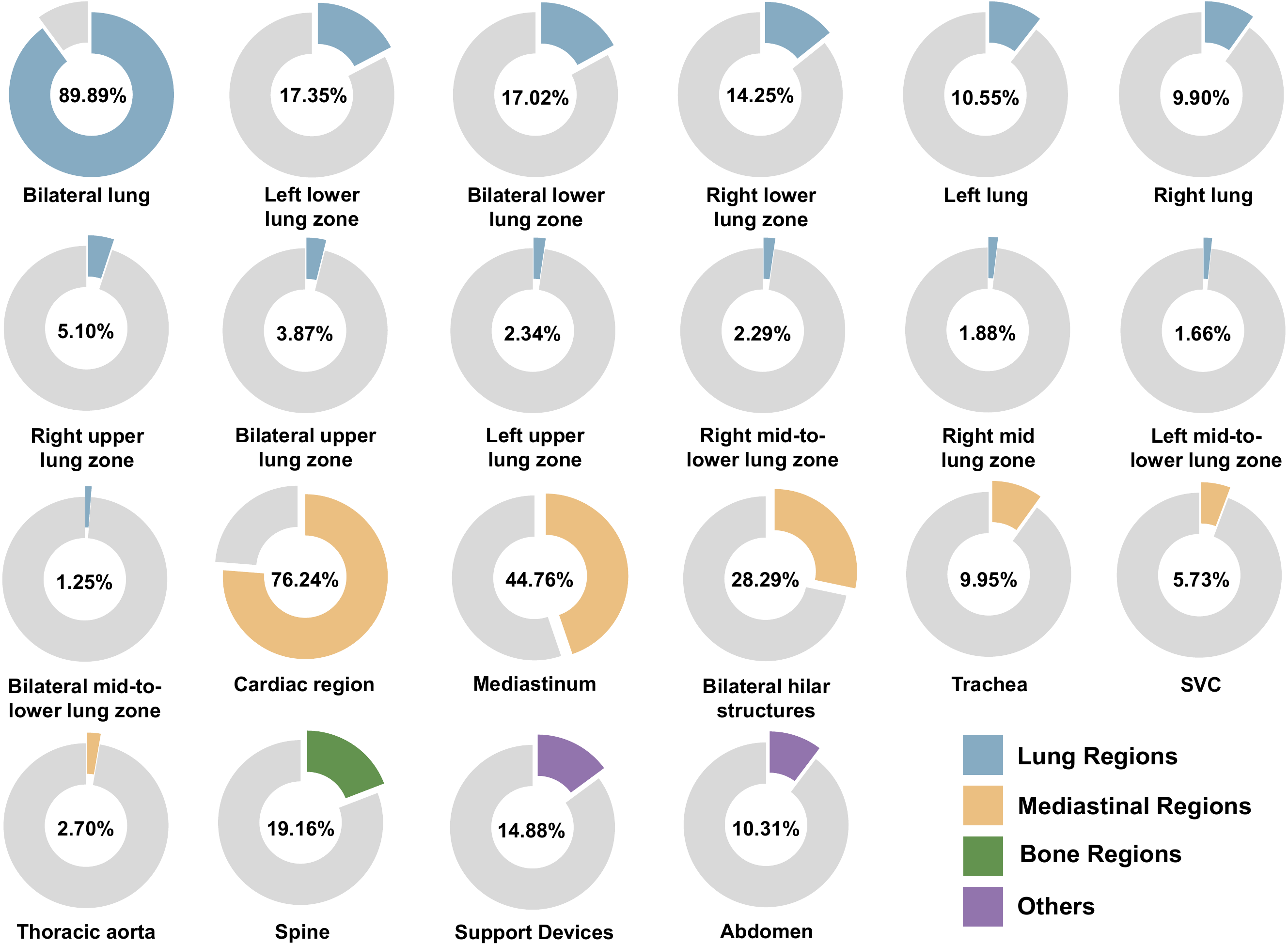}
    \label{regiondistr}
\end{figure*}

We further report the distribution of fine-grained anatomical regions in MMRad-22K. As shown in Figure~\ref{regiondistr}, the dataset exhibits a clearly long-tailed regional distribution, with \textit{lung} and \textit{mediastinal} related regions accounting for the largest proportion of evidence, broadly consistent with the dominant focus of CXR interpretation. At the same time, less frequent but clinically relevant regions from the \textit{bone} and \textit{others} groups are also represented, including spine, abdomen, and support-device-related regions. This suggests that the coarse anatomy-guided grouping used in MMRad-22K still preserves substantial fine-grained regional diversity.

\subsection{More Information about Evaluation Metrics}
\label{evaluation_metrics}

\begin{itemize}
    \item \textbf{BLEU} \citep{papineni2002bleu} measures n-gram precision between the generated report and the reference report.
    \item \textbf{METEOR} \citep{banerjee2005meteor} combines precision, recall, and a fragmentation penalty, while additionally considering word order and synonym matching.
    \item \textbf{ROUGE} \citep{lin2004rouge} measures n-gram recall with respect to the reference report.
    \item \textbf{RadGraph F1} \citep{delbrouck2022improving} extracts radiology entities and relations and computes an F1 score over their overlap, mainly for chest X-ray reports.
    \item \textbf{RaTEScore} \citep{zhao2024ratescore} evaluates radiology reports at the entity level by matching clinically important entities with type-aware embedding similarity, making it robust to synonyms and sensitive to negation.
    \item \textbf{RadCliQ} \citep{yu2023evaluating} combines BLEU, BERTScore, CheXbert vector similarity, and RadGraph F1 into a composite score that is optimized to better align with radiologist judgment.
\end{itemize}

\subsection{More Information About Evidence Construction and Inference Settings}
\label{evidence_settings}

Each anatomy-guided evidence unit in MMRad-22K contains three aligned components: a regional textual observation, an anatomical bounding box, and a localized image crop. By training on these paired multimodal evidence structures together with the final report target, the adapted Anole model acquires both localization-related prediction ability and visual evidence generation ability within a unified autoregressive framework.

This design gives rise to two practical ways of incorporating localized visual evidence at inference time. In the \textbf{Generated} setting, we follow Anole's native multimodal generation paradigm and use autoregressively generated visual evidence together with regional textual observations. In the \textbf{Grounded} setting, the model first predicts anatomical regions, which are then mapped back to the source chest X-ray to extract localized image crops as grounded visual evidence, which is aligned with the crop-based localized evidence setting examined in Section~\ref{benefit}.

\subsection{More Cases Generated by Anole-MMRad}
\label{cases}

\subsubsection{Representative Cases}

\begin{figure*}[t]
    \centering
    \caption{\textbf{Representative cases of report generation with structured multimodal evidence.} Each example shows the full chest X-ray image, regional evidence units, the predicted report, and the reference report. The cases cover both relatively normal and abnormal studies, illustrating that structured multimodal evidence can support clinically coherent study-level report generation. Green shading and blue shading represent consistent descriptions for the lung and mediastinal region.}
    \includegraphics[width=\textwidth]{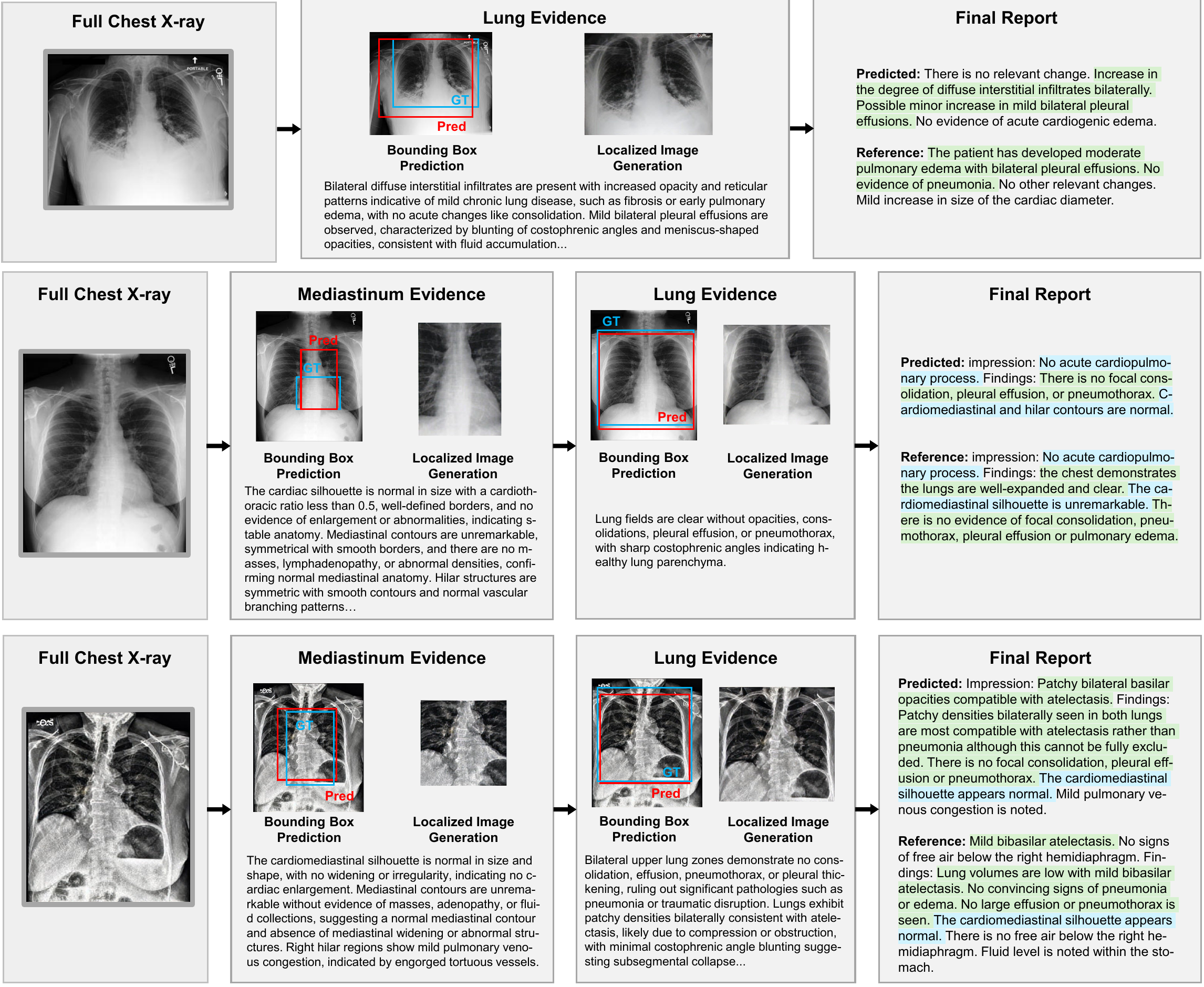}
    \label{moregenerated}
\end{figure*}

Figure~\ref{moregenerated} presents representative examples of report generation with structured multimodal evidence. The examples cover both normal and abnormal studies, illustrating that the structured regional evidence can support clinically coherent study-level reports across different levels of difficulty. In relatively normal cases, the model produces findings that closely match the reference reports. In abnormal cases, it is able to combine multiple localized evidence units to capture major findings such as pulmonary edema, pleural effusion, or bibasal opacities, demonstrating the benefit of anatomy-guided multimodal evidence for report generation.

\subsubsection{Failure Cases}

\begin{figure*}[t]
    \centering
    \caption{\textbf{Failure cases of report generation with structured multimodal evidence.} Each example shows the full chest X-ray image, regional evidence units, the predicted report, and the reference report. Although the overall study-level impression remains broadly reasonable, these cases show that some supported findings may still be under-emphasized or omitted in the final report. Green shading, blue shading, and orange shading represent consistent descriptions for the lung, mediastinal, and bone region. Red shading represents errors.}
    \includegraphics[width=\textwidth]{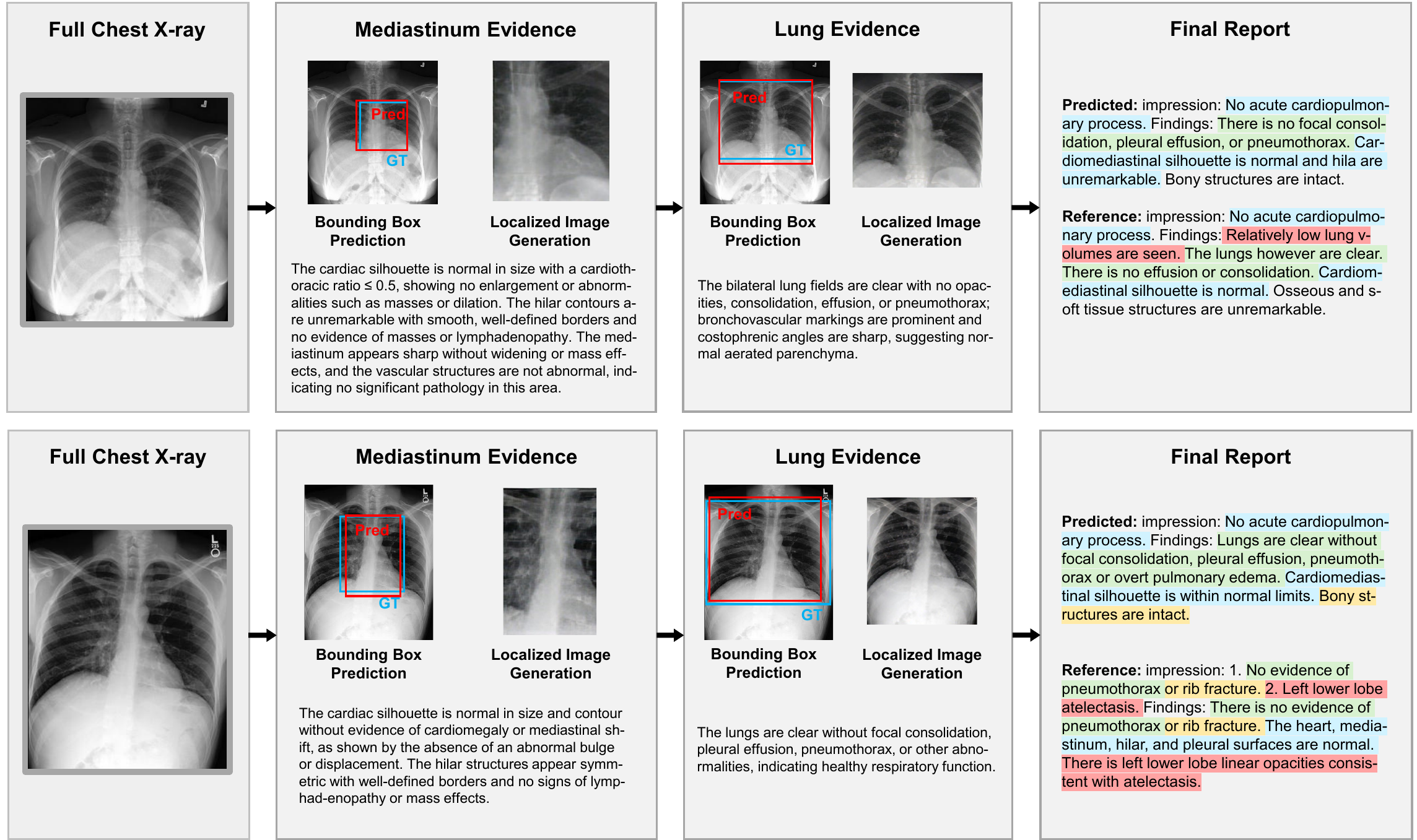}
    \label{failure}
\end{figure*}

Figure~\ref{failure} presents representative failure cases. In both examples, the model retains a broadly reasonable study-level impression but does not fully capture all findings described in the reference report. The first case omits low lung volumes, whereas the second misses a more localized left lower lobe linear opacity/atelectatic change. These cases suggest that, even with anatomy-guided multimodal evidence, some supported findings may be under-emphasized or omitted when they are integrated into the final study-level report.

\subsection{Ethics, Data Use, and Disclosure Information}

\subsubsection{Potential Risks.}
MMRad-22K is constructed for research on chest X-ray report generation and multimodal evidence modeling. Potential risks include propagation of source reporting biases, under-representation of rare findings or patient subgroups, and inappropriate over-trust in automatically generated reports. In addition, because the dataset reorganizes report-aligned supervision into study-level evidence, generated outputs may appear clinically plausible even when they omit important details. For these reasons, MMRad-22K and models trained on it should be used only in research settings and not for direct clinical decision-making.

\subsubsection{Data Sources and Usage Terms.}
MMRad-22K is derived from MIMIC-CXR~\cite{johnson2019mimic} and GEMeX-ThinkVG~\cite{liu2025gemexthink}, both of which are existing research resources with their own access conditions and usage requirements. Our use of these resources is limited to research purposes and follows their intended access setting. MMRad-22K is released only as a derived research artifact consistent with the original access restrictions, and it is not intended to circumvent the access controls, licensing conditions, or redistribution boundaries of the underlying data sources.

\subsubsection{Intended Use and Distribution Boundary.}
The intended use of MMRad-22K is academic research on chest X-ray report generation, multimodal evidence organization, and related evaluation. It is not intended for clinical deployment, patient care, or autonomous medical reporting. Any released derivative artifact should remain compatible with the original data-access conditions, and the dataset should not be used outside research environments where these conditions do not apply.

\subsubsection{Use of AI Assistants.}
Large language models were used during dataset construction to reformulate question-level grounded traces into report-generation-oriented evidence units and to support intermediate verification, as described in Section~\ref{dataconstruction}. These model outputs were not accepted without control: they were subjected to multi-stage verification, including self-checking, report-level consistency checking, and clinician adjudication for uncertain cases. AI assistants were also used for limited writing support during manuscript preparation, with all technical content, experimental claims, and final wording reviewed and revised by the authors.

\end{document}